\documentclass[letterpaper, 10 pt, conference]{ieeeconf}  

\IEEEoverridecommandlockouts                              

\usepackage{amsmath} 
\usepackage{amssymb}  

\usepackage{graphics} 
\usepackage{epsfig} 
\usepackage{hyperref}
\usepackage{cleveref}
\crefname{figure}{Fig.}{Figs.}
\Crefname{figure}{Fig.}{Figs.}
\crefname{table}{Table}{Tables}
\Crefname{table}{Table}{Tables}
\usepackage{mathrsfs}
\usepackage{tabularx}
\usepackage{booktabs}
\usepackage{subcaption}

\title{\LARGE \bf
Effect of Temporal Resolution on Zero-Shot Semantic \\
Understanding of Rapid Human Actions
}

\author{
Yongpeng Cao$^{1, *}$\thanks{$^{*}$Corresponding author}%
\thanks{$^{1}$Institute of Industrial Science, The University of Tokyo, Japan}
 and Yuji Yamakawa$^{2}$%
\thanks{$^{2}$Interfaculty Initiative in Information Studies, Graduate School of Interdisciplinary Information Studies, The University of Tokyo, Japan}%
\\ \{cao, y-ymkw\}@iis.u-tokyo.ac.jp
}

\begin{document}
\bstctlcite{IEEEtranBSTcontrol}

\maketitle
\thispagestyle{empty}
\pagestyle{empty}

\begin{abstract}
Understanding human actions from visual observations is essential for human--robot interaction, particularly when semantic interpretation of unfamiliar or hard-to-annotate actions is required. For rapid or less common activities, collecting sufficient labeled data for supervised learning is challenging, making zero-shot approaches a practical alternative for semantic understanding without task-specific training. While recent advances in large-scale pre-trained models enable such zero-shot reasoning, the impact of temporal resolution, especially for rapid and fine-grained motions, remains underexplored.

In this study, we investigate temporal-resolution conditions for zero-shot semantic understanding of high-speed human actions. Using kendo as a representative case of rapid and subtle motion patterns, we propose a training-free pipeline that combines a pre-trained video-language model for semantic representation with large language model-based reasoning for pairwise action comparison. Our primary experiment evaluates full-observation raw videos from 10 to 120 Hz in 10-Hz increments. Nearest class prototype (NCP) accuracy is highest at 120 Hz, and the mean over 70--120 Hz is higher than that over 10--60 Hz. The corresponding mean prototype margin changes from -5.74 to +2.52 similarity-score points between these two halves and is also highest at 120 Hz. Secondary experiments show that tracking overlays reduce mean full-observation performance but improve mean partial-observation performance. These findings provide preliminary evidence that retaining high-temporal-resolution visual information can benefit training-free semantic understanding of rapid human actions.

\end{abstract}

\section{Introduction}

Understanding human actions from visual observations is a fundamental problem in human--robot interaction (HRI), particularly for applications requiring timely and proactive responses. In real-world scenarios such as collaborative tasks or sports interactions, robots must interpret human actions from incomplete and rapidly evolving motion cues. This challenge is further amplified for high-speed actions, where critical discriminative information may occur within very short temporal intervals.

Recent advances in action recognition have been largely driven by supervised learning methods, including skeleton-based graph convolutional networks and video-based recognition models. While these approaches achieve strong performance, they heavily rely on task-specific training data and often assume standard video frame rates (e.g., 30 FPS). In contrast, zero-shot and training-free approaches, enabled by vision-language models (VLMs) and large language models (LLMs), provide a promising alternative for action understanding without requiring labeled datasets. However, existing work primarily focuses on general daily actions and overlooks the impact of temporal resolution, especially for fast and fine-grained motions.

In this work, we investigate how temporal resolution affects zero-shot semantic discriminability of human actions. Specifically, we select high-speed kendo attack actions as a representative case of rapid and subtle motion patterns and examine whether their semantic differences can be captured without any task-specific training. We propose a training-free pipeline that leverages a pre-trained video-language model to generate textual descriptions from short video clips, followed by an LLM that reasons about and measures pairwise action similarity.
We evaluate every 10-Hz condition from 10 to 120 Hz. The results show that the 120 Hz input produces stronger semantic separability than the temporally downsampled inputs under the proposed fixed-frame semantic encoding setting. To further reflect practical human--robot interaction scenarios, we additionally evaluate the proposed framework under partial observation conditions for early action understanding, where only the initial stage of an action is available before completion. Furthermore, we analyze the role of tracking-based human joint information and its interaction with temporal resolution under both full and partial observation settings. A preliminary version of this study was presented at a workshop~\cite{cao2026highspeed}, in which the effect of temporal resolution was evaluated at 30, 60, and 120 Hz. The present work extends it through a systematic evaluation from 10 to 120 Hz at 10-Hz increments and a prototype margin analysis across different conditions. This dense frame-rate analysis is intended to characterize the sensitivity of semantic understanding in the proposed pipeline to temporal resolution, investigate potential trends across frame rates, and examine whether an effective frame-rate range can be identified for the task. Furthermore, the prototype margin analysis provides a measure of semantic separability that may not be captured by NCP accuracy alone. The main contributions of this work are summarized as follows:
\begin{itemize}
    \item We extend the earlier three-frame-rate study to a systematic 10--120 Hz evaluation at 10-Hz increments under full and partial observation, with and without tracking overlays.
    \item We add a prototype-margin analysis to quantify semantic separation beyond binary NCP accuracy in our training-free VLM--LLM pipeline.
    \item We show that full/raw NCP accuracy and prototype margin are stronger overall in the upper half of the frequency range and peak at 120 Hz; secondary results reveal different tracking effects under full and partial observation.
\end{itemize}

\section{Related Work}

\subsection{Human Action Recognition}

Human action recognition plays an important role in HRI. Conventional approaches commonly rely on skeleton-based representations and deep learning architectures such as Graph Convolutional Networks (GCNs) to capture spatiotemporal motion patterns from human body joints~\cite{yan2018spatial, CTR-GCN, cheng2020skeleton, degcn, blockgcn, protoGCN, chi2022infogcn, 2sagcn2019cvpr, liu2023temporal}. These methods have demonstrated strong performance on common action recognition benchmarks. However, most existing approaches are training-based and require large amounts of annotated data, which limits their applicability to new, rare or highly specialized actions.

\subsection{Vision--Language Models for Action Understanding}

Recent advances in VLMs and LLMs have enabled semantic understanding without task-specific training. MotionCLIP~\cite{motionclip} projects human motion into the CLIP~\cite{CLIP} embedding space, enabling semantic alignment between motion and text. Zhang et al.~\cite{zhang2024pevl} utilize pre-trained VLMs for contrastive learning, which enhances the alignment between visual features and textual semantics. In addition, LLM-based reasoning methods have recently been adopted for action recognition. LLM-AR~\cite{qu2024llms} converts human motions into linguistic descriptions and formulates action recognition as a language understanding problem. Similarly, GAP~\cite{xiang2023generative} utilizes an LLM to generate semantic descriptions of movements and incorporates these descriptions into action recognition.

Building upon these foundations, further research has extended these models to training-free zero-shot action understanding. These approaches leverage semantic reasoning and pre-trained knowledge to recognize actions. FreeZAD~\cite{freezad} performs action localization and recognition directly, demonstrating that pre-trained VLMs can generalize to unseen actions through semantic matching. Similarly, TEAR~\cite{TEAR} enhances action understanding through action descriptors and predicts actions through similarity matching between textual and visual embeddings.

Despite these advances, most VLM-based action understanding studies focus on daily activities captured at standard frame rates. The effect of temporal resolution on semantic understanding of human actions remains largely underexplored, particularly for rapid and fine-grained movements such as martial arts actions.

\subsection{Early Action Recognition under Partial Observation}

Early action recognition aims to identify human actions before the complete motion sequence becomes available. This capability is essential for real-time robotic systems, where response latency directly affects interaction quality and safety. Existing methods typically perform action prediction from partial observations. Stergiou and Damen~\cite{stergiou2023wisdom} use attention to weight discriminative early motion frames, while~\cite{liu2023rich} models action-semantic consistent knowledge. However, these approaches generally rely on supervised training and large annotated datasets. As a secondary analysis complementing the full-observation, this work explores whether partial observations of high-speed actions preserve sufficient semantic information for zero-shot understanding. Our experiments further analyze how temporal resolution influences semantic separability under incomplete observation conditions.

\subsection{Kendo Motion Analysis and High-Speed Perception}

Several studies have investigated human--robot interaction and motion recognition in kendo. The sword-fighting robot proposed in~\cite{Namiki2015} estimates the human motion trajectory and generates corresponding defensive actions. Kendo action recognition has also been explored using wearable sensors together with the LSTM method~\cite{strikethrust2019}. Moreover, our previous work~\cite{kendo_cao2022} explores markerless high-speed kendo action tracking and recognition.

Different from these training-based or rule-based approaches, this work interprets high-speed kendo actions in a training-free setting. Instead of relying on handcrafted criteria or wearable sensors, we leverage vision-language models and large language models to perform semantic comparison.

\section{Methodology}

\begin{figure}[htbp!]
\centering
    \rotatebox{90}{\includegraphics[height=0.86\linewidth]{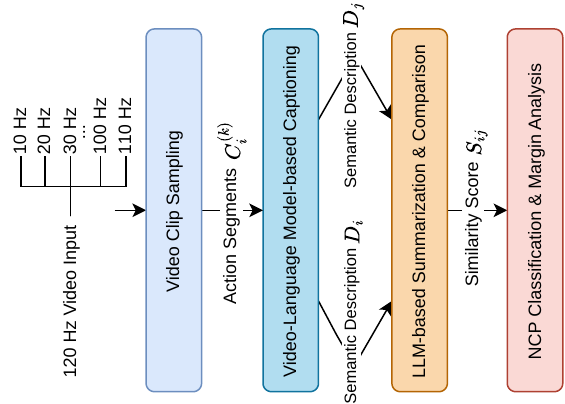}}
\caption{System framework.}
\label{fig:sys_framework}
\end{figure}

In this work, we propose a training-free semantic comparison pipeline to investigate whether high-speed kendo actions can be semantically interpreted without task-specific training. \Cref{fig:sys_framework} shows the overall pipeline. We explore whether semantic discriminability between different attack patterns can be preserved in a zero-shot, training-free setting. The goal of this proof-of-concept experiment is to examine the feasibility of zero-shot semantic understanding of human actions and whether higher temporal resolution supports more stable semantic extraction of rapid motion.

Each sample video is first processed using a fixed-length clip segmentation strategy. Specifically, a video $V_i$ is processed as a sequence of clips with a temporal window of $L$ frames. The $k$-th clip is defined as:
\begin{equation}
C_i^{(k)} = \{ f_{i,t} \mid t = (k-1)L + 1, \dots, kL \},
\end{equation}
where $f_{i,t}$ denotes the $t$-th frame of video $V_i$. This results in a set of clips $\{ C_i^{(k)} \}_{k=1}^{K_i}$ that capture local temporal dynamics of the action. $K_i$ here denotes the total number of clips for video $V_i$, which depends on the video length and the chosen clip length $L$. When fewer than $L$ frames remain, the final clip is shifted backward and may overlap the preceding clip.
Each clip is processed by a pre-trained video-language model to generate a textual description:
\begin{equation}
d_i^{(k)} = \mathrm{VLM}(C_i^{(k)}),
\end{equation}
where $d_i^{(k)}$ denotes the description corresponding to the $k$-th clip. Each video $V_i$ is therefore represented as a sequence of textual descriptions:
\begin{equation}
D_i = \{ d_i^{(k)} \}_{k=1}^{K_i}.
\end{equation}

To measure semantic similarity between actions, we adopt a pairwise comparison strategy using an LLM. For each pair of videos $(i, j)$, the LLM takes the corresponding sequences of descriptions $(D_i, D_j)$ as input and assigns a similarity score to each action pair following an LLM-as-a-judge paradigm, reflecting their semantic proximity:
\begin{equation}
S_{ij} = \mathrm{LLM}(D_i, D_j),
\end{equation}
where $\mathrm{LLM}(\cdot, \cdot)$ denotes the 0--100 similarity score generated based on the comparison of the two description sequences. The resulting similarity matrix $\mathbf{S} \in [0,100]^{N \times N}$ captures pairwise semantic relationships across all samples.

Finally, to quantify whether the generated semantic descriptions preserve class-discriminative information, we adopt a leave-one-out Nearest-Class Prototype (NCP) evaluation~\cite{NCP}. For each query sequence, its average semantic similarity to the remaining examples of each action category is computed, and the category with the highest average similarity is selected as the prediction. The query sample itself is excluded when constructing the prototype of its ground-truth class. Although no task-specific training is performed for semantic representation generation, this evaluation uses the known class grouping of the remaining samples. For each sample $i$ and class $c$, we compute:
\begin{equation}
\mu_i^{(c)} = \frac{1}{|\mathcal{I}_c\setminus\{i\}|}
\sum_{j \in \mathcal{I}_c\setminus\{i\}} S_{ij},
\end{equation}
where $\mathcal{I}_c$ denotes the index set of samples in class $c$. The predicted label for sample $i$ is then given by:
\begin{equation}
\hat{y}_i = \arg\max_c \ \mu_i^{(c)}.
\end{equation}
If two or more classes have the same highest score, the prediction is regarded as a failed classification.

We also measure how far each sample is from the strongest incorrect prototype. The sample margin and its mean are
\begin{equation}
m_i=\mu_i^{(y_i)}-\max_{c\ne y_i}\mu_i^{(c)},\qquad
M=\frac{1}{N}\sum_{i=1}^{N}m_i.
\end{equation}
A positive margin means that the ground-truth prototype has the highest similarity. This metric retains information discarded by binary NCP accuracy.

Additionally, to evaluate the proposed framework under real-time human--robot interaction conditions, we add an early action recognition setting using partial observations. Instead of using the full action sequence, only the initial stage of each action is retained according to the segmentation criterion introduced in our previous work~\cite{kendo_cao2022}. This setting simulates practical scenarios where robotic systems must infer human intent before the action is fully completed in order to enable timely responses.

\section{Experiments}

This proof-of-concept study examines temporal-resolution sensitivity in a training-free VLM--LLM pipeline for semantic information extraction of rapid human actions. The main experiment is the 10--120 Hz full/raw baseline condition; tracking-overlay and partial-observation conditions are secondary comparisons to the baseline. 

\subsection{High-Speed Video Collection and Preprocessing}
To investigate the effect of temporal resolution on zero-shot semantic understanding, we collected high-speed video data of kendo attack actions. Videos were captured at 120 Hz using a monochrome high-speed camera, with action durations ranging from 2 to 3 seconds. We use the Ximea MQ013MG-ON monochrome camera, which provides high frame rates for capturing fine-grained motion details at sufficient resolution.

Each video was temporally downsampled to produce comparison conditions from 10 to 120 Hz in 10-Hz increments.

\begin{figure}[htbp!]
\centering
\begin{subfigure}{0.47\linewidth}
    \includegraphics[width=\linewidth]{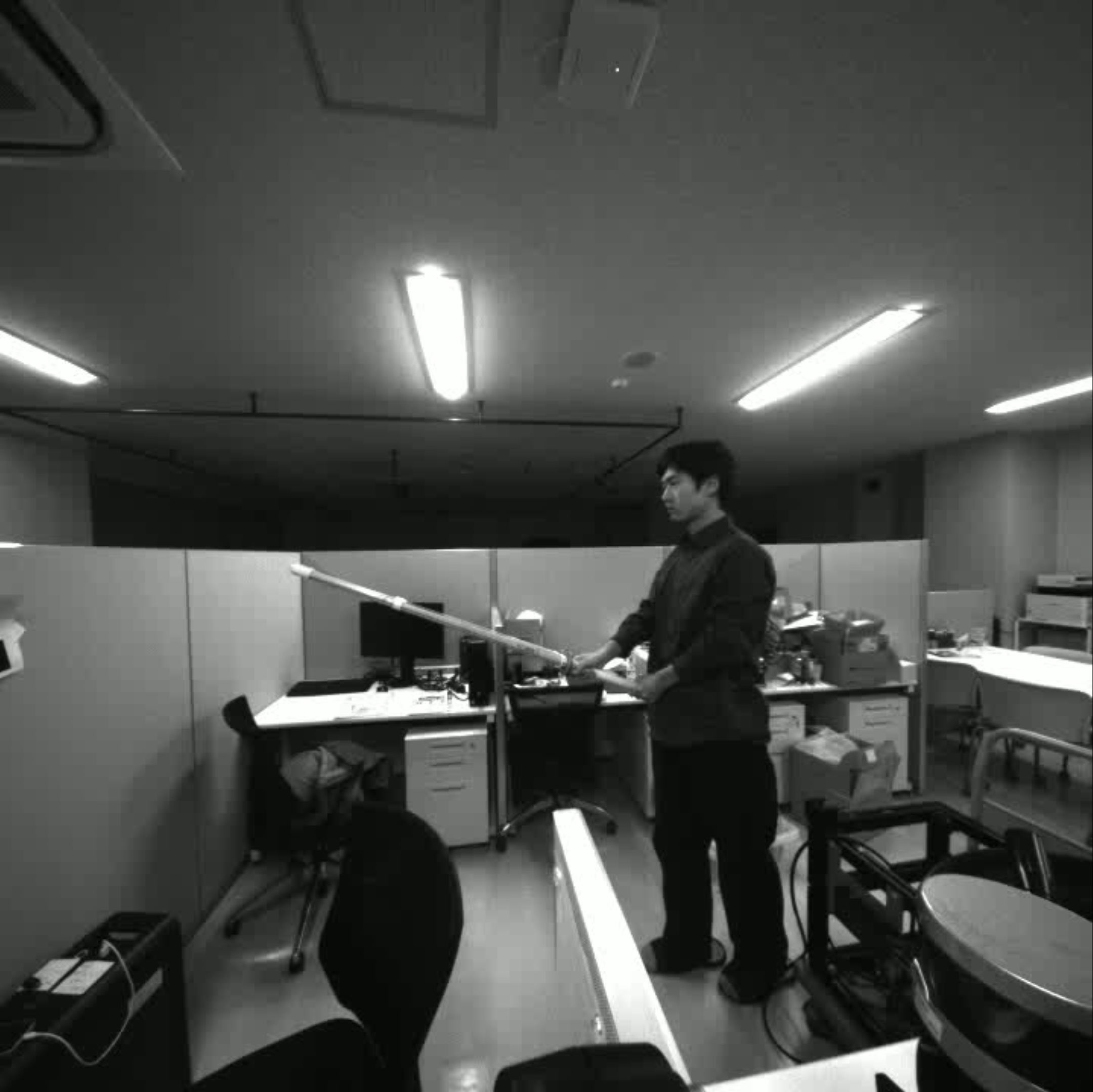}
    \caption{Without tracking overlay}
    \label{fig:no_tracking}
\end{subfigure}
\hfill
\begin{subfigure}{0.47\linewidth}
    \includegraphics[width=\linewidth]{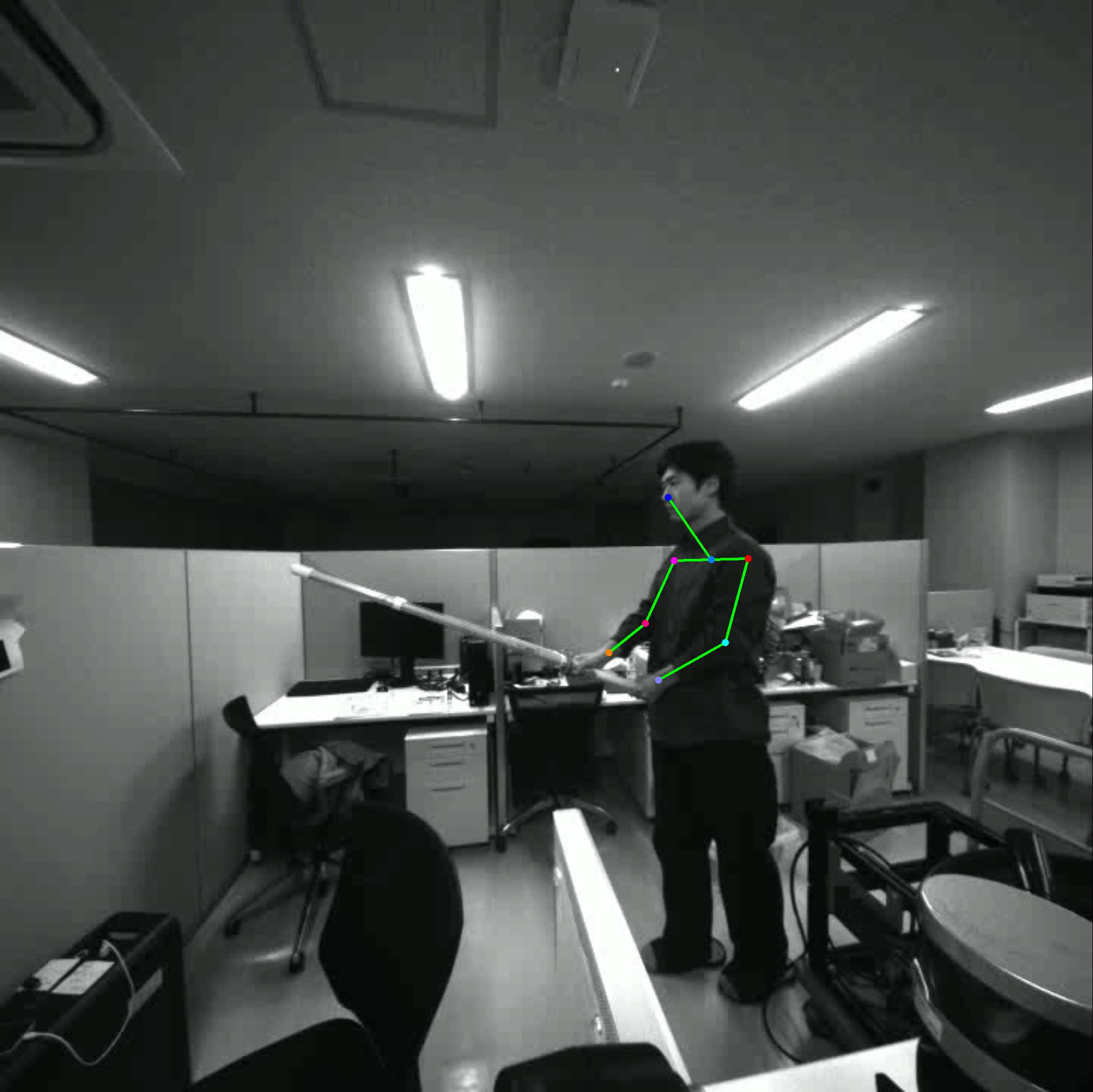}
    \caption{With tracking overlay}
    \label{fig:with_tracking}
\end{subfigure}
\caption{Example high-speed kendo video frames without and with a joint-tracking overlay. Reproduced from~\cite{cao2026highspeed}.}
\label{fig:tracking_overlay}
\end{figure}

In addition, we applied a pose estimation model to extract upper-body joint positions, which were overlaid on the original video frames to examine the impact of human joint information on semantic understanding. We adopted the joint-tracking approach proposed in our previous work~\cite{kendo_cao2022} to obtain smooth joint trajectories. The tracking results were visualized as colored skeletons, with different colors representing different joints. \Cref{fig:tracking_overlay} shows an example of the tracking results overlaid on the original video frames.

\subsection{Data Preparation}

We prepare three trials for each attack pattern: Men, Kote, and Dou, together with a blank category without an attack action to serve as a baseline for comparison. For clarity, we denote the action labels as M (Men), K (Kote), D (Dou), and R (Regular), with trial indices from 1 to 3. Original kendo action videos were captured at 120 Hz.

First, video-level semantic representations are generated from action frames. InternVideo2.5~\cite{wang2025internvideo2} is adopted to generate semantic summaries from short video clips. In the experiments, we group 16 consecutive frames into one chunk to produce a semantic description.

In these experiments, \textit{Qwen3-4B-Instruct-2507} is selected as the pre-trained LLM for both reasoning and comparison. The model is prompted to focus on temporal ordering, attack patterns, target regions, and key motion transitions, while no task-specific fine-tuning is applied.

\subsection{Results}
The definition of ``the same action'' is inherently ambiguous in martial arts movements. Even actions belonging to the same category may differ in execution details or temporal transitions. Therefore, the primary objective of this experiment is semantic comparison rather than strict action classification. We use NCP accuracy only to quantify the class discriminability preserved by the generated semantic descriptions.

\subsubsection{Effect of Temporal Resolution}
\begin{figure}[!htbp]
\centering
    \includegraphics[width=0.98\linewidth]{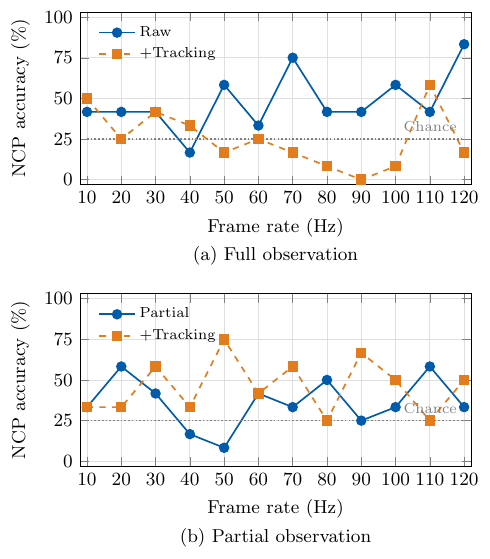}
\caption{NCP accuracy from 10 to 120 Hz under (a) full and (b) partial observation. Lines with round markers denote the raw condition; dashed lines and square markers denote the tracking-overlay condition.}
\label{fig:fps_accuracy}
\end{figure}

\Cref{fig:fps_accuracy} plots accuracy across every evaluated frequency under different conditions, and \Cref{tab:fps_results} reports the values. Primary full/raw accuracy reaches 83.3\% at 120 Hz. Its mean over 70--120 Hz is 57.0\%, compared with 38.9\% over 10--60 Hz. The points remain non-monotonic. Thus, the primary results support a broad higher-frame-rate advantage but not a monotonic relationship or precise saturation frame rate.

\begin{figure}[!htbp]
\centering
    \includegraphics[width=0.98\linewidth]{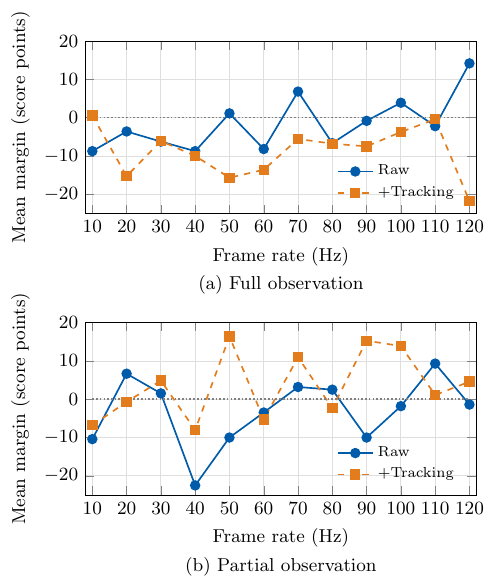}
\caption{Mean prototype margin from 10 to 120 Hz under (a) full and (b) partial observation. Lines with round markers denote the raw condition; dashed lines and square markers denote the tracking-overlay condition.}
\label{fig:prototype_margin}
\end{figure}

\begin{table}[!t]
\caption{NCP Accuracy (\%) Across Evaluated Conditions.}
\label{tab:fps_results}
\centering
\normalsize
\setlength{\tabcolsep}{4pt}
\begin{tabular}{@{}ccccc@{}}
\toprule
\shortstack{Frame rate\\(Hz)} & \shortstack{Full\\raw} & \shortstack{Full\\+track.} & \shortstack{Partial\\raw} & \shortstack{Partial\\+track.} \\
\midrule
10  & 41.7 & 50.0 & 33.3 & 33.3 \\
20  & 41.7 & 25.0 & 58.3 & 33.3 \\
30  & 41.7 & 41.7 & 41.7 & 58.3 \\
40  & 16.7 & 33.3 & 16.7 & 33.3 \\
50  & 58.3 & 16.7 &  8.3 & 75.0 \\
60  & 33.3 & 25.0 & 41.7 & 41.7 \\
70  & 75.0 & 16.7 & 33.3 & 58.3 \\
80  & 41.7 &  8.3 & 50.0 & 25.0 \\
90  & 41.7 &  0.0 & 25.0 & 66.7 \\
100 & 58.3 &  8.3 & 33.3 & 50.0 \\
110 & 41.7 & 58.3 & 58.3 & 25.0 \\
120 & 83.3 & 16.7 & 33.3 & 50.0 \\
\bottomrule
\end{tabular}
\end{table}

\begin{figure*}[!htbp]
\centering
    \includegraphics[width=0.99\textwidth]{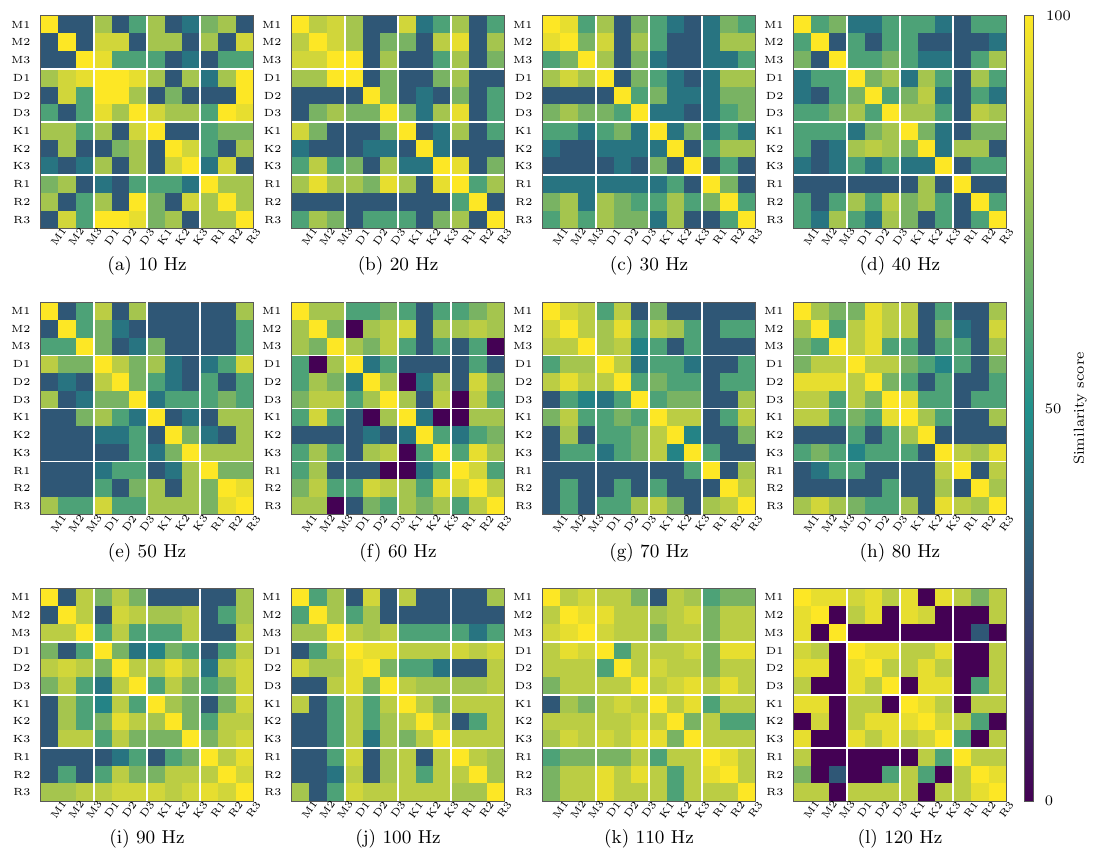}
\caption{Full/raw pairwise similarity matrices at all evaluated frame rates from 10 to 120 Hz in 10-Hz increments. Samples are ordered by class (M: Men, D: Dou, K: Kote, R: Regular).}
\label{fig:similarity_matrices}
\end{figure*}
\subsubsection{Semantic Separation Beyond Accuracy}
The continuous prototype-margin curves in \Cref{fig:prototype_margin} further support the same interpretation. For the primary full/raw condition, mean margin changes from $-5.74$ similarity-score points over 10--60 Hz to $+2.52$ over 70--120 Hz and reaches its maximum of $+14.17$ at 120 Hz. The curve remains non-monotonic, but the shift from a negative to a positive half-range mean indicates stronger semantic separation in the upper half. The paired curves, compared with the full observation condition, also show the positive effects of the tracking overlay on partial observation.

\subsubsection{Effect of Tracking Information}
The aggregate comparison in \Cref{tab:tracking_summary} shows that tracking overlays have different effects for the two observation settings. For full observations, tracking reduces mean accuracy from 47.9\% to 25.0\%, and changes the overall mean margin from $-1.61$ to $-8.83$. For partial observations, tracking increases mean accuracy from 36.1\% to 45.8\%, wins at 7 of the 12 frame rates, and changes the overall mean margin from $-3.03$ to $+3.66$. Thus, both metrics indicate that the overlay generally degrades full-observation performance but often benefits partial observation in this experiment.

\begin{table}[!htbp]
\caption{Mean Effect of Tracking Across Frame Rates. Wins, ties, and losses compare tracking overlay with the raw condition.}
\label{tab:tracking_summary}
\centering
\normalsize
\setlength{\tabcolsep}{2.8pt}
\begin{tabular}{@{}lcccc@{}}
\toprule
Obs. & \shortstack{No track.\\mean (\%)} & \shortstack{+Track.\\mean (\%)} & \shortstack{Difference\\(pp)} & \shortstack{Tracking\\win/tie/loss} \\
\midrule
Full    & 47.9 & 25.0 & $-22.9$ & 3/1/8 \\
Partial & 36.1 & 45.8 & $+9.7$  & 7/2/3 \\
\bottomrule
\end{tabular}
\end{table}
\subsubsection{Representative Similarity Structure}
\Cref{fig:similarity_matrices} provides qualitative context for the primary full/raw result at all evaluated frame rates from 10 to 120 Hz in 10-Hz increments. At 120 Hz, the Dou, Kote, and Regular blocks contain strong within-class regions. Therefore, the matrices illustrate both the 120 Hz maximum accuracy and its remaining errors.

\subsection{Discussion}

The experimental results show that the primary full/raw condition has a higher mean in the 70--120 Hz half of the full range, with the 120 Hz condition achieving the highest accuracy. The prototype margin measures each sample’s separation from its strongest incorrect class and shows the same upper-half frequency advantage. Together, these observations are consistent with a benefit from retaining fine temporal information.

Retaining more temporal samples may preserve critical transitions that help distinguish rapid and similar actions. Unlike training-based models that learn task-specific discriminative features, the proposed pipeline relies on VLM captions and LLM-based comparison, which may be sensitive to sparsely sampled motion cues. Higher temporal resolution introduces greater fault tolerance, making the overall semantic comparison more robust for rapid movements.

Moreover, tracking overlays are not clearly beneficial for complete action recognition, but improve the semantic separation for partial observations. This suggests that tracking information may be useful during key action periods, while it may be redundant or distracting when overlaid across the entire action sequence.

There are several limitations to this work. Given its proof-of-concept nature, the evaluation uses a limited number of action instances and a relatively simple semantic comparison pipeline. The early stopping of recording requires predefined criteria for actions, which limits the generalizability to online inference. The recognition accuracy is also limited by human pose estimation accuracy, and similarity scores between different actions tend to remain generally high across all frame rates. This reflects the generalization tendency of LLM-based reasoning, which prioritizes overall semantic resemblance over fine-grained execution differences. Larger-scale evaluation, more robust semantic scoring, and a systematic analysis that controls observation duration and semantic window size are left for future work.

\section{Conclusion}
This proof-of-concept study investigates temporal-resolution conditions for zero-shot semantic understanding of high-speed human actions, using kendo as a representative case. We propose a training-free pipeline that combines video-language models and LLM-based reasoning to compare action pairs without task-specific training, with evaluation on 12 frame rates and the prototype-margin analysis. NCP accuracy and prototype margin are highest at 120 Hz and stronger on average over 70--120 Hz than over 10--60 Hz in the primary condition. Secondary comparisons show that tracking generally reduces full-observation performance but more often improves partial observation. These findings provide preliminary evidence that high-temporal-resolution input can be useful for semantic reasoning about rapid human actions. Future work will investigate larger datasets, more robust semantic scoring, reasoning strategies, and improved multimodal integration.

\bibliographystyle{IEEEtran}
\bibliography{main}

\end{document}